\documentclass[10pt,journal,compsoc, number]{IEEEtran}
\ifCLASSOPTIONcompsoc
  \usepackage[nocompress]{cite}
\else
  \usepackage{cite}
\fi
\ifCLASSINFOpdf
\else
\fi

\usepackage{comment}

\usepackage{cite}
\usepackage{amsmath,amssymb,amsfonts}
\usepackage{algorithmic}
\usepackage{graphicx}
\usepackage{textcomp}
\usepackage{xcolor}

\usepackage{listings}
\usepackage{booktabs}
\usepackage{multirow}
\usepackage{subfigure}
\usepackage{caption}

\newtheorem{defn}{Definition}[section]

\begin{document}
%
\title{
PKGM: A Pre-trained Knowledge Graph Model for E-commerce Application
}

%
%

\author{Wen Zhang$^1$,
Chi-Man Wong$^5$,
Ganqiang Ye$^2$, 
Bo Wen$^1$,
Hongting Zhou$^2$,
Wei Zhang$^5$,
Huajun Chen$^{2,3,4}$ \\ 
\textit{\small{$^1$School of Software Technology, Zhejiang University}} \\
\textit{\small{$^2$College of Computer Science and Technology, Zhejiang University}} \\
\textit{\small{$^3$Alibaba-Zhejiang University Joint Research Institute of Frontier Technologies}}\\
\textit{\small{$^4$Hangzhou Innovation Center, Zhejiang University}}\\
\textit{\small{$^5$Alibaba Group}}
}

\markboth{Journal of \LaTeX\ Class Files,~Vol.~14, No.~8, August~2015}%
{Shell \MakeLowercase{\textit{et al.}}: Bare Demo of IEEEtran.cls for Computer Society Journals}
%



\IEEEtitleabstractindextext{%
\begin{abstract}
In recent years, knowledge graphs have been widely applied as a uniform way to organize data and have enhanced many tasks requiring knowledge. In online shopping platform Taobao, we built a billion-scale e-commerce product knowledge graph. It organizes data uniformly and provides item knowledge services for various tasks such as item recommendation. Usually, such knowledge services are provided through triple data, while this implementation includes (1) tedious data selection works on product knowledge graph and (2) task model designing works to infuse those triples knowledge. More importantly, product knowledge graph is far from complete, resulting error propagation to knowledge enhanced tasks. To avoid these problems, we propose a Pre-trained Knowledge Graph Model (PKGM) for the billion-scale product knowledge graph. On the one hand, it could provide item knowledge services in a uniform way with service vectors for embedding-based and item-knowledge-related task models without accessing triple data. On the other hand, it's service is provided based on implicitly completed product knowledge graph, overcoming the common the incomplete issue. We also propose two general ways to integrate the service vectors from PKGM into downstream task models. We test PKGM in five knowledge-related tasks, item classification, item resolution, item recommendation, scene detection and sequential recommendation. Experimental results show that  PKGM introduces significant performance gains on these tasks, illustrating the useful of service vectors from PKGM.  
\end{abstract}

\begin{IEEEkeywords}
Knowledge Graph, Pre-training, Embedding, E-commerce Application. 
\end{IEEEkeywords}}

\maketitle

\IEEEdisplaynontitleabstractindextext

%
\IEEEpeerreviewmaketitle

\IEEEraisesectionheading{\section{Introduction}
\label{sec:introduction}}
\IEEEPARstart{O}{nline} shopping has greatly contributed to the convenience of people's life and  
the e-commerce era witnessed rapid development in the past a few years. The growing transaction on the e-commerce platform is based on billions of items of all kinds.
Those items should be well organized to support the daily business. 
Due to the convenience of fusing data from various sources and building semantic connections between entities, knowledge graphs(KG) is usually applied. It represent facts as triples, such as \emph{(iPhone, brandIs, Apple)}. 
We built an e-commerce Product Knowledge Graph (PKG) and make it a uniform way to integrate massive information about items on Taobao platform.  Currently, PKG contains 70+ billion triples and 3+ million rules.
It greatly contributing to item knowledge services. For example, 
It help provide item knowledge services and support a variety of item-knowledge-related tasks including searching, question answering, recommendation, and business intelligence, etc.



To serve for these item-knowledge-related tasks, the common paradigm is providing 
triples related to target items 
in PKG to task managers. 
With this paradigm, there are tedious works for data selection and knowledge-enhanced model designing. 
More importantly,   
similar to other KGs, PKG is still far from complete. 
Triple data with key information missing fed into these tasks may bias or even mislead them. 
Thus we are seeking for alternative ways for PKG to provide knowledge service for items.   

Since the concept of `pre-training and fine-tuning' has proven to be very useful in the Natural Language Processing (NLP) community \cite{BERT}, which refers to pre-training a language model in a huge amount of text and then fine-tune it on downstream tasks such as sentiment analysis \cite{ACL2020_Pretrain4SA}, relation extraction \cite{ACL2019_Pretrain4RE} and text classification \cite{EMNLP2019_Pretrain4TC} with a small amount of data.
Thus we are inspired to consider pre-training PKG and make it possible to conveniently and effectively serve for item related tasks in vector space.
The `convenience' refers to provide knowledge service for items in a uniform way, and downstream tasks do not need to revise their model much to adapt to item knowledge, where tedious model design works are avoided. 
The `effectiveness' refers to that 
the knowledge services from PKG 
overcome the incompleteness issue of PKG and  those tasks enhanced with knowledge services could achieve better performance, especially with a small amount of data. 
Thus our research question is 
\emph{whether we could pre-train PKG and make it a knowledge provider for item-knowledge-related tasks, which could avoid tedious data selection and model design, and overcome the incompleteness of PKG.}


Similar to Pre-trained Language Models (PLMs), 
The purpose of \textbf{P}re-trained \textbf{K}nowledge \textbf{G}raph \textbf{M}odel (\textbf{PKGM}) is to learn entity and relation embeddings in continuous vector space and provide knowledge services downstream task through which they could get necessary fact knowledge via calculation with embeddings without accessing triple data.

With the paradigm of providing triple data, 
there are two kinds of common queries executed:
(1) triple queries querying the tail entity or property value of the an item given a relation or property,
(2) relation queries querying whether one item has a given relation or attribute.
Considering the incompleteness issue of  PKG, the PKGM should be capable of  
(1) showing or predicting what is the tail entity for a given entity and relation,
(2) showing whether a relation exists for an entity. 


Thus our proposed PKGM includes two modules. One is \textbf{Triple Query Module} encoding the truth value of an input triple, which could serve for triple queries after training. Since a lot of knowledge graph embedding methods are proposed for triple encoding and link prediction, we apply the simple and effective TransE \cite{TransE-Bordes-2013} in the triple query module.
The other one is \textbf{Relation Query Module} encoding the existence of a relation for a given entity, which could serve for the relation query after training. 
In relation query module, we 
learn an entity transformation matrix $\mathbf{M}_r$ for each relation $r$. With $\mathbf{M}_r$, we make the transformed embedding of head entity $\mathbf{h}$ approaches to relation embedding $\mathbf{r}$ if $h$ owns relation $r$. 

After pre-training, the triple query module and the relation query module  provide knowledge service vectors  given an entity.
More specifically, triple query module provides service vectors with tail entity embeddings given a target entity. For example, suppose that the target entity is a smartphone, the triple query module will provide the predicted tail entity embedding of key relations including \emph{brandIs, seriesIs, memoryIs} and so on, no matter triples about these relations of the target entity exists in PKG or not. 
Relation query module provides service vectors indicating the  existence of  relations for target entities, and  service vectors will approach to zero vector if the target entity has or should have input relations. 
Combining service vectors from the triple query and relation query module, PKGM could provide knowledge services to enhance other tasks with item knowledge contained in  PKG without accessing triple data. 
 
With service vectors, we propose general ways to incorporate them in embedding-based methods for different tasks. We classify embedding-based methods into two types according to the number of input embeddings for target entities. One is methods with a sequence of embeddings for a target entities, and the other one with a single embedding.
For sequence-embedding inputs, we extend the input sequence by appending service vectors after original embeddings for target entities. And for single-embedding input, we condense service vectors into one and concatenate it with the single embedding of target entities. 

We pre-train PKGM on the billion-scale PKG  and  enhance five item-related tasks with service vectors from PKGM including item classification, item resolution, item recommendation, scene detection and sequential recommendation, following the proposed general ways for servicing.
Experimental results show that PKGM successfully improves the performance of these five tasks, especially with a small amount of training data. 

In summary, contributions of this work are as follows: 
\begin{itemize}
    \item We propose a way to pre-train a knowledge graph, which could provide knowledge services vectors to enhance other tasks in a general way. 
    \item We propose a PKGM with two significant advantages, one is completion capability, and the other one is triple data independency.
    \item We practice PKGM on billion-scale product knowledge graph and test it on five item related tasks, showing that PKGM successfully enhances them with item knowledge and improves their performance, especially with a small amount of training data.
\end{itemize}

\section{Preliminary}
\subsection{Product Knowledge Graph}
Alibaba has accumulated one hundred billion scale of product data in Product Knowledge Graph (PKG), contributed from various markets (Taobao, Tmall, 1688, AliExpress, etc.), brand manufacturers, industry operations, governance operations, consumers, state agencies and so on. 
For providing better shopping experience, standardization of product (e.g., normalization of product standard or completeness of product information) and deep connection mining between internal and external data, are essential to E-commerce businesses. 
Advanced NLP methods, semantic reasoning, and deep learning methods can be developed for various tasks such as  search, recommendation, platform governance, question answering, brand manufacturers. 
Currently, PKG contains four key components, standard products, standard brands, standard barcodes and standard categories. They are developed from nine major ontologies such as public sentiments, encyclopedia, and national industry standards, and so on, via entity recognition, entity linking and semantic analysis.

Similar to other knowledge graphs, 
PKG stores information of items and relationships between them as triples,  represented as  $\mathcal{K}_P = \{\mathcal{E}, \mathcal{R}, \mathcal{T}\}$, 
where $\mathcal{E}, \mathcal{R}$ and $\mathcal{T}$ are entity, relation and triple set respectively. 
$\mathcal{R} = \{ \mathcal{P} \cup \mathcal{R^\prime} \}$ 
is composed by a set of items' properties $\mathcal{P}$ and a set of relationships between items $\mathcal{R}^\prime$. 
$\mathcal{E} = \{\mathcal{I}\cup \mathcal{V}\}$ contains a set of items $\mathcal{I}$ and a set of values $\mathcal{V}$. 
Currently, PKG contains more than $700$ billion triples. 
\subsection{Knowledge Services}
\label{sec:knowledge-services}
Knowledge services that PKG provides for other tasks refers to returning data in PKG that matches input queries. 
There are two types of queries commonly executed in PKG, triple query and relation query. 

\textbf{Triple Query}
is to get the tail entity given an head entity $h$ and relation $r$. 
The stored tail entities will be returned if there are triples stored about $h$ with $r$ as relations, otherwise nothing will be returned. 

\textbf{Relation Query}
is to get the relations existing for a given entity $h$.
As a result, relations that participants in triples with $h$ as head entity will be returned.  

Triple query (left) and relation query (right) formed in  SPARQL are as follows:
\begin{lstlisting}[frame=shadowbox]
SELECT ?x                SELECT ?x
WHERE {h r ?x}           WHERE {h ?x ?y}
\end{lstlisting}
The whole view of PKG could be built by combining these two types of queries. 

\section{Pre-trained Knowledge Graph Model}

\begin{figure*}
    \centering
    \includegraphics[scale=0.4]{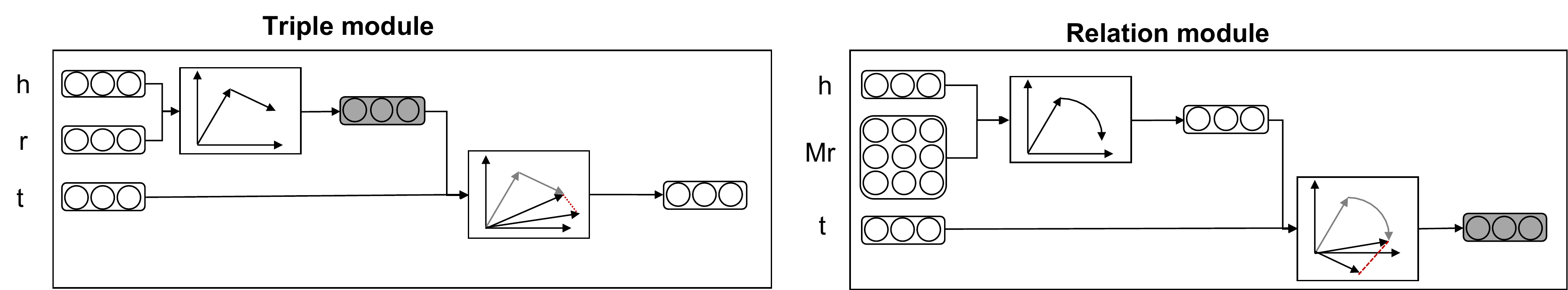}
    \caption{Pre-trained Knowledge Graph Model.}
    \label{fig:PKGM}
\end{figure*}

We propose a \textbf{P}re-trained \textbf{K}nowledge \textbf{G}raph \textbf{M}odel (\textbf{PKGM}) to represent and complete $\mathcal{K}_P$ in continuous vector space, based on which item knowledge services could be provided for other tasks through calculation in vector space. There are three steps for in PKGM:
\begin{itemize}
    \item \textbf{Pre-training}: Firstly, we pre-train PKGM on billion-scale PKG, making it gain the capability of providing knowledge services for triple and relation queries. 
    \item \textbf{Servicing}: Secondly, for a target item that knowledge is required from other tasks, PKGM provides service vectors containing item information.
    \item \textbf{Applying}: Thirdly, with service vectors from PKGM, we inject them into embedding-based models for item-knowledge-related tasks in one of the two general ways that we proposed. 
\end{itemize}


\subsection{Pre-training}
In order to make it possible to access all possible triples in PKG through PKGM, based on analysis about knowledge services in Section \ref{sec:knowledge-services}, we build two query modules in PKGM to simulate the knowledge accessing in continuous vector space. 
One is \textbf{Triple Query Module} and the other one is \textbf{Relation Query Module}, as shown in Fig~\ref{fig:PKGM}.

\subsubsection{Triple query module $\mathcal{M}_{triple}$}
For a triple query $Q_{triple}(h, r)$, $\mathcal{M}_{triple}$ will provide a service vector indicating the candidate tail entity. 

To gaining such capability, 
we make an assumption in  $\mathcal{M}_{triple}$ for triple $(h,r,t)$, that the head entity $h$ and relation $r$ could be transferred to tail entity $t$ in vector space if $(h,r,t)$ is true. We encode such assumption via a score function $f_{triple}(h, r, t)$, the result of which  will encode the truth value of $(h,r,t)$.

A lot of knowledge graph embedding methods\cite{KGSurvey} have been proposed to encode the truth value of triples and proved to be effective. 
Considering the extremely large scale of PKG, we apply TransE \cite{TransE-Bordes-2013} with translation assumption in triple query module due to its simplicity and effectiveness. 
We encode each $e \in \mathcal{E}$ and $r \in \mathcal{R}$ as a vector, called embeddings and  represented with characters in bold. Following the translation assumption, for each positive triple $(h, r, t)$, we make $\mathbf{h} + \mathbf{r} \approx \mathbf{t}$, where $\mathbf{h} \in \mathbb{R}^d, \mathbf{r}\in \mathbb{R}^d$ and $\mathbf{t}\in \mathbb{R}^d$ are embedding of $h, r$ and $t$ respectively. Thus the score function for $(h, r, t)$ is 
\begin{equation}
f_{triple}(h,r,t) = \|\mathbf{h} + \mathbf{r} - \mathbf{t}\|  
\label{score-triple}
\end{equation}
where $\| \mathbf{x} \|$ is the L1 norm of vector $\mathbf{x}$. 
During training, we make $\mathbf{h} + \mathbf{r} \approx \mathbf{t}$ for positive triple existing in PKG and $\mathbf{h} + \mathbf{r} $ far away from $\mathbf{t}$ for negative ones. 

\subsubsection{Relation query module $\mathcal{M}_{rel}$}
For a relation query $Q_{rel}(h,r)$, $\mathcal{M}_{rel}$ will pride a service vector indicating whether there is or should be triples in PKG with $h$ as head entity and $r$ as relation. 

To gain such capability, we make an assumption following triple query module. For head entity and relation pair $(h, r)$, the $\mathbf{h}$ could be transferred to $\mathbf{r}$ if $(h,r)$ exists. 
We encode such assumption via a score function $f_{rel}(h,r)$, the result of which will encode the truth value of pair $(h,r)$. 

In vector space, we make zero vector $\mathbf{0}$ represent EXISTING, thus  $f_{rel}(\mathbf{h}, \mathbf{r}) \approx \mathbf{0}$ if $(h,r)$ exists, and
$f_{rel}(\mathbf{h}, \mathbf{r})$ far away from $\mathbf{0}$ otherwise. 
We design a transferring matrix for each relation $r$, denoted as $\mathbf{M}_r \in \mathbb{R}^{d \times d}$. 
$\mathbf{M}_r$ helps transfer $\mathbf{h}$ to relation space via multiplication between them.
Thus $f_{rel}$ is designed as follow
\begin{equation}
f_{rel}(h, r) = \| \mathbf{M}_r \mathbf{h} - \mathbf{r}\|
\label{score-rel}
\end{equation}
For a positive pair $(h,r)$, $f_{rel}(h,r)$ should be as small as possible, and as large as possible for negative pairs. 

\subsubsection{Loss function}
Foe a triple $(h,r,t)$, with score from $f_{triple}(h,r,t)$ and $f_{rel}(h,r)$, we take the summation of them as the final score 
\begin{equation}
    f(h,r,t) = f_{triple}(h,r,t) + f_{rel}(h,r)
\end{equation}
in which positive triples should have a small score and and negative triples should have a large one. 
Following previous work\cite{TransE-Bordes-2013}, we make a margin-based loss as training objective 
\begin{equation}
L = \sum_{(h,r,t) \in \mathcal{K}}  [f(h,r,t) + \gamma - f(h',r',t') ]_+  
\label{loss}
\end{equation}
where $(h',r',t')$ is a negative triple generated for $(h,r,t)$ by randomly sample an entity $e \in \mathcal{E}$ to replace $h$ or $t$, or randomly sample a relation $r' \in \mathcal{R}$ to replace $r$. $\gamma$ is a hyperparameter representing the margin should be achieved between positive sample scores and negative sample scores. 
\begin{equation}
[x, y]_+=\left\{
\begin{aligned}
x &  \text{\; if \; } x \ge y \\
y & \text{\; if \; } x < y 
\end{aligned}
\right.
\end{equation}

\subsection{Servicing}

\subsubsection{Service for triple queries $\mathcal{S}_{triple}(h,r)$}
Given a triple query $\mathcal{Q}_{triple}(h,r)$,
PKGM provides the candidate tail entity in vector space via returning its embedding
\begin{equation}
    S_{triple}(h,r) = \mathbf{h} + \mathbf{r} 
\end{equation}
with which the out put of $S_{triple}(h,r)$  will approximate to $\mathbf{t}$ if $(h,r,t) \in \mathcal{K}_{P}$ as a result of training objective (Equation(\ref{loss})). If there is no triple in $\mathcal{K}_P$ with $h$ as head entity and $r$ as relation, the out put of  $S_{triple}(h,r)$  will be an entity representation that is most likely to be the right tail entity, as a results of the widely proved and inherent capability of triple completion of KGE  methods\cite{KGSurvey}.

\subsubsection{Service for relation queries $\mathcal{S}_{rel}(h,r)$}
Given a relation query $\mathcal{Q}_{rel}(h,r)$, PKGM provides the existence of head entity and relation pair $(h,r)$ in vector space via returning a vector denoting the existence
\begin{equation}
    S_{rel}(h, r) = \mathbf{M}_r \mathbf{h} - \mathbf{r} 
\end{equation}
For the output of $S_{rel}(h, r)$, 
there are three kinds of situations: 
(1) If there is triple in $\mathcal{K}_P$ with $h$ as head entity and $r$ as relation,   $S_{rel}(h, r)$ will approximate to EXIST embedding $\mathbf{0}$,
(2) If there is no triple in $\mathcal{K}_P$ in the form of $(h,r,e)$ where $e\in \mathcal{E}$, but there should be, $S_{rel}(h, r)$ will also approximate to EXIST embedding $\mathbf{0}$. This is the relation completion capability of PKGM gaining from Equation(\ref{score-rel}). 
(3) If there is no triple in $\mathcal{K}_P$ in the form of $(h,r,e)$, and there should not be, $S_{rel}(h, r)$ will be away from  $\mathbf{0}$. 

For more concise understanding, 
we summarize the function for pre-training and servicing in Table~\ref{tab:PKGM-functs}.
\begin{table}[!hbpt]
    \centering
    \caption{Functions for pre-training and servicing of PKGM. The subscript letter $T$ and $R$ denotes for $triple$ and $rel$.}
    \begin{tabular}{c|c | c}
    \toprule
        \textbf{Module} & \textbf{Pre-training} & \textbf{Servicing} \\
         \midrule
        Triple Query
        & $f_{T}(h,r,t) = \|\mathbf{h} + \mathbf{r} - \mathbf{t}\|$    
        & $S_{T}(h,r) = \mathbf{h} + \mathbf{r}$ \\
        \midrule
        Relation Query 
        & $f_{R}(h, r) = \| \mathbf{M}_r \mathbf{h} - \mathbf{r}\| $ 
        & $S_{R}(h, r) = \mathbf{M}_r \mathbf{h} - \mathbf{r}$\\
        \bottomrule
    \end{tabular}
    \label{tab:PKGM-functs}
\end{table}

\subsubsection{Advantages of PKGM service}
There are three significant advantages  of getting item knowledge  via PKGM's servicing given an entity $h$ and a relation $r$:
\begin{itemize}
\item We could access the tail entity in an implicit way via calculation in vector space without truly querying triples existing in PKG. This makes query service independent to data and ensures data privacy. 
\item Results for each input pairs are uniformed as service vectors from two query modules instead of triple data,  making it easier to inject them into downstream task models since designing model to encode triple data is avoided.  
\item We could get the inferred tail entity $t$ even there is no triple $(h,r,t) \in \mathcal{K}_P$, which greatly overcomes the incompleteness disadvantages of PKG.  
\end{itemize}

\subsection{Applying}
Given an item as target entity $e$,
several service vectors from triple query module will be given, denoted as  $\mathbf{S}_{triple}^e = [\textbf{S}_1^e, \textbf{S}_2^e, ..., \textbf{S}_k^e]$ and also service vectors from relation query module, denoted by $\textbf{S}_{rel}^e = [\textbf{S}_{k+1}^e, \textbf{S}_{k+2}^e, ..., \textbf{S}_{2k}^e]$. $k$ is the number of \textbf{key relations} for $e$ generated according PKG. Key relations $\mathcal{R}_{e}$ are those relations important to $e$.

In order to applying those service vectors in other item-knowledge-related tasks, we propose two general ways to integrate them into embedding-based models. 
According to the number of embeddings for an target entity in inputs of a model, we categorize embedding-based models into two classes, one is \textbf{sequence-embedding models} inputting a sequence of  embeddings, and the other one is \textbf{single-embedding models} inputting a single embedding.  

\subsubsection{For Sequence-embedding Models}
\begin{figure}[!hbpt]
    \centering
    \includegraphics[width=0.35\textwidth]{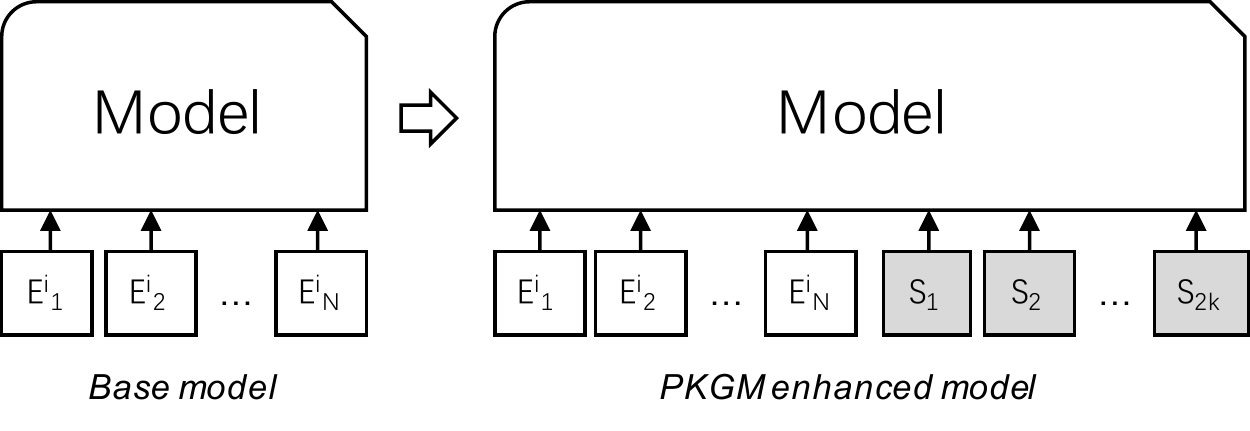}
    \caption{The key idea of applying service vectors into  sequence-embedding models.}
    \label{fig:PKGM+1}
\end{figure}

Sequence-embedding models refers to those models containing a sequence embeddings inputs for a target entity $e$. 
In these models,
the sequential embeddings are usually generated according to target item's  side information, like word embeddings of description text or labeled feature embeddings. 
We generalize such model in left part in Fig.\ref{fig:PKGM+1} as base model and represent the sequence input as $\textbf{E}^e = [\textbf{E}_1^e, \textbf{E}_2^e, ..., \textbf{E}_N^e]$. 

Considering that those sequential models are able to automatically deal with inputs of different lengths, we propose to append all service vectors included in $\mathbf{S}_{triple}$ and $\mathbf{S}_{rel}$ for $e$  at the end of $\mathbf{E}^e$. 
After appending the service vectors, the input will be  $\hat{\textbf{E}^e}= [\textbf{E}_1^e, \textbf{E}_2^e, ..., \textbf{E}_N^e, \textbf{S}_1^e, \textbf{S}_2^e, ..., \textbf{S}_{2k}^e]$, where we firstly append $\textbf{S}^e_{triple}$ and then append $\textbf{S}^e_{rel}$ by default.
After extend the input sequence, the embedding-based model will make service vector interact with original inputs automatically.
We show the key idea of integrating service vectors into sequence-embedding models in Fig.\ref{fig:PKGM+1}. 

\subsubsection{For Single-embedding Models}
\begin{figure}[!htbp]
    \centering
    \includegraphics[width=0.4\textwidth]{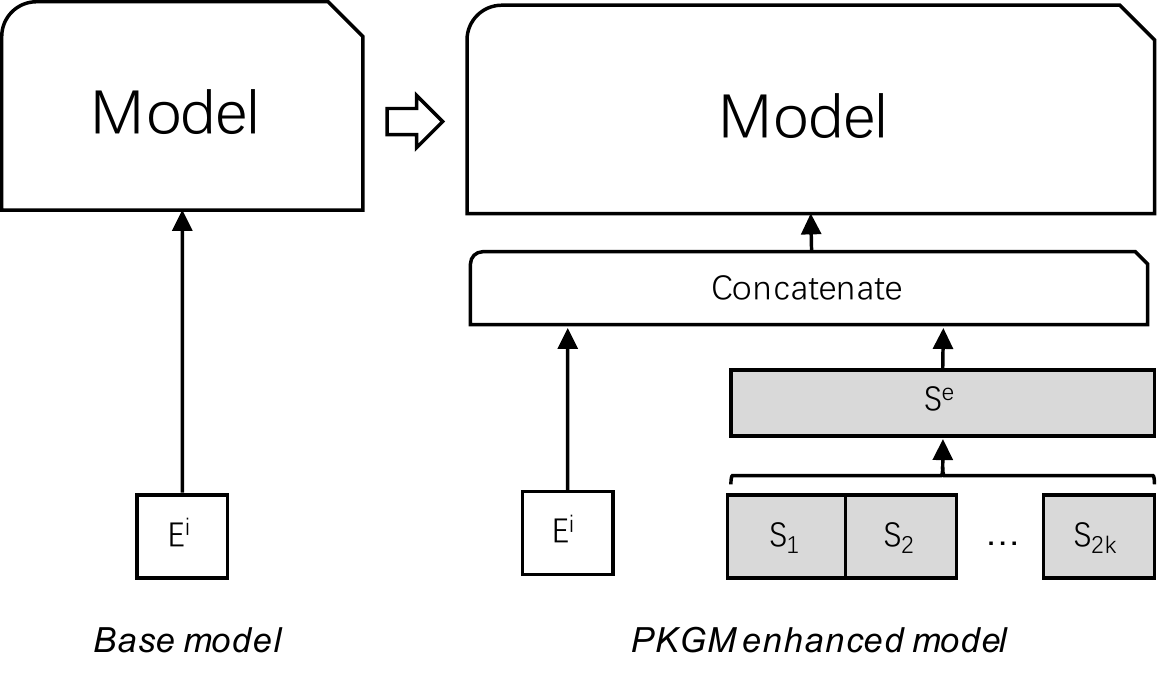}
    \caption{The key idea of applying service vectors into  single-embedding models.}
    \label{fig:PKGM+2}
\end{figure}

Single-embedding models refer to those models containing only one input embedding for a target entity $e$. 
In these models, 
the single embedding usually refers to the embedding of $e$ in current latent vector space and is learnt during training. We represent the single embedding as $\mathbf{E}^e$. We show the general overview of single-embedding model in the left part of Fig.\ref{fig:PKGM+2}.


Since there are only one embedding for $e$ in the base model, we propose to firstly combine $\mathbf{S}_{triple}^e$ and $\mathbf{S}_{rel}^e$ into one, represented as $S^e$ and then concatenate it with original item embedding $\textbf{E}^e$ to replace $\textbf{E}^e$ as input, as shown in Fig.\ref{fig:PKGM+2}.

During combination, there are two ways recommended considering base model's and PKGM's embedding dimension. If their embedding dimension are comparable, combination is recommended as follow
\begin{equation}
    S^e = \frac{1}{k}\sum_{i \in [1, k]} \hat{S_i}^e
\end{equation}
\begin{equation}
    \hat{S_i}^e = [S_i^e;S_{i + k}^e], \text{where}\; i \in [1, k])
\end{equation}
where $[x;y]$ means concatenation of  vector $x$ and $y$. 
If PKGM's embedding dimension is significantly smaller than base model, combination is recommended as \begin{equation}
    S^e = [\mathbf{S}_1^e, \mathbf{S}_2^e, ..., \textbf{S}_{2k}^e]
\end{equation}
Besides above two ways, other ways of combination also could be  applied.


\section{Experiments}
In this section, we firstly introduce the details of pre-training PKGM on a billion-scale PKG, and introduce four servicing types that PKGM could provide, among which one is baseline with item embedding and the other three are based on service vectors. Then we introduce the details of experiments on  5 knowledge-enhanced item-related tasks with PKGM, including item classification, item resolution, item recommendation, scene detection and sequential recommendation. For each task, we test PKGM on 3 datasets to explore how would PKGM help downstream tasks with  different scales. 

\subsection{Pre-training}
\textbf{Dataset.}
We pre-train PKGM on the billion scale PKG. 
For pre-training, we remove relations with occurrences less than 5000 in PKG, since they are likely noisy and contain little information that not only increase the model's parameters but also might deteriorate pre-training results. 
Statistic details of the dataset for pre-training are shown in Table\ref{tab:dataset-pre-train}.   
\begin{table}[!hbpt]
    \centering
    \caption{Statistics of dataset for pre-training.}
    \begin{tabular}{  c|c | c}
    \toprule
             \textbf{\# Entity} & \textbf{\# Relation} & \textbf{\# Triples} \\
           \midrule 
           142,641,094 &426 &1,366,109,966
 \\
        \bottomrule
    \end{tabular}
    \label{tab:dataset-pre-train}
\end{table}

\noindent \textbf{Training details.}
We implement PKGM with popular large-scale machine learning package Tensorflow~\cite{tensorflow} and Alibaba's  Graph-learn\footnote{https://github.com/alibaba/graph-learn}.
Graph-learn~\cite{zhu2019aligraph} is a large-scale distributed framework for node and edge sampling in graph neural network (GNN). We use Graph-learn to perform edge sampling, and 1 negative triple is sampled for each triple.  
During training, we adopt Adam\cite{Adam} with initial learning rate as 0.0001 for model optimization and make batch size as  1000.  We set the entity and relation embedding dimension $d$ to $64$ as a balance between model's servicing capability and storage requirement for billion-scale PKG. The margin in loss function $\gamma$ is set to $1$. 
Finally, the model size is 88GB. We train it  with 50 parameter servers and 200 workers for 2 epochs. The whole pre-traing step took about 15 hours.

\noindent \textbf{Key relation generation.}
The service provided by PKGM is related to key relations for target items. 
Thus for each item $e$, we select $10$ key relations for it according relational frequency based on category which is defined as follows,
\begin{equation}
    f(r, e) = \sum _{(e^\prime, isA, c(e))\in PKG} q(e^\prime, r)
\end{equation}
where $c(e)$ is the category of $e$ which means $(e, isA, c(e)) \in \mathcal{K}_p$ and  
$$q(e^\prime, r)=
\begin{cases}
1& \exists (e^\prime, r, X) \in \mathcal{K}_p \\
0& \nexists (e^\prime, r, X) \in \mathcal{K}_p  \\
\end{cases}$$
More specifically, suppose $e$ belongs to category $c(e)$, we gather all items belonging to $c(e)$ and account the frequency of relations that those items have with other entities. Finally, we select top $10$ most frequent ones as key relations for $e$. 

\subsection{Servicing}
During servicing for downstream tasks, we explore one baseline service with item embeddings and three variations of providing service vectors from PKGM. 
\begin{itemize}
    \item \textbf{PKGM-item} is a baseline that provide single item embedding from PKGM for one item in downstream tasks. 
    \item \textbf{PKGM-all} provides $2 \times k$ service vectors composed of $k$ from triple query module and $k$ from relation query module for one item in downstream tasks. 
    \item \textbf{PKGM-T} provides only $k$ service vectors from triple query module for one item in downstream tasks. 
    \item \textbf{PKGM-R} provides only $k$ service vectors from relation query module for one item in downstream tasks. 
\end{itemize}
During downstream task training, vectors from PKGM are fixed that will not be updated.
\subsection{TASK1: Item Classification}

\subsubsection{Task definition}
The target of item classification is to assign an item to a class in the given class list. 
Usually, item titles are used for classification since most items in our platform having a title filled by sellers. We frame item classification as a text classification task as follows
\begin{defn}
(Item classification)  Given a set of data $\mathcal{D} = \{ \mathcal{P}, \mathcal{T}, \mathcal{C}, \mathcal{R} \}$, where $\mathcal{P}, \mathcal{T}$  and $\mathcal{C}$ are a set of product, titles and classes respectively,  $\mathcal{R} = \{(p, t, c) | p \in \mathcal{P},t \in \mathcal{T} \text{and}
\; c \in \mathcal{C})\}$ is a record set of the class and title of each product. Each title $t = [ w_1, w_2, w_3, ..., w_n ]$ is an ordered sequence composed by words, the target is to train a mapping function $f: \mathcal{T} \mapsto \mathcal{C}$. 
\end{defn}

\begin{table*}[!hbpt]
    \centering
    \caption{Results for item classification task}
    \resizebox{\textwidth}{!}{
    \begin{tabular}{l|c c c | c c c | c c c }
    \toprule
    \multirow{2}{*}{\textbf{Method}}  & \multicolumn{3}{c|}{\textbf{ItemCla-large}} &  \multicolumn{3}{c|}{\textbf{ItemCla-medium}} &  \multicolumn{3}{c}{\textbf{ItemCla-small}} \\ \cline{2-10}
    & Hit@1 & Hit@3 & Hit@10  & Hit@1 & Hit@3 & Hit@10 & Hit@1 & Hit@3 & Hit@10 \\
    \midrule
      BERT  
      & .710(0.00\%) & .849(0.00\%) & .925(0.00\%) 
      & .610(0.00\%) & .790(0.00\%) & .892(0.00\%)
      & .303(0.00\%) & .470(0.00\%) & .681(0.00\%) 
      \\
      BERT$_{PKGM-item}$ 
      & .713($\uparrow${0.42\%}) 
      & .851($\uparrow${0.24\%}) 
      & .928($\uparrow${0.32\%})
      
      & .611($\uparrow${0.16\%})
      & .796($\uparrow${0.76\%})
      & .902($\uparrow${1.12\%})
      
      & .305($\uparrow${0.66\%}) 
      & .475($\uparrow${1.06\%}) 
      & .690($\uparrow${1.32\%})
      \\
       \midrule
      BERT$_{PKGM-T}$   
      & .713($\uparrow${0.42\%}) 
      & .858($\uparrow${1.06\%}) 
      & .931($\uparrow${0.65\%})
      
      & \textbf{.617}($\uparrow${1.15\%}) 
      & .790(\textcolor{black}{0.00\%}) 
      & .901($\uparrow${1.01\%})
      
      & .307($\uparrow${1.32\%}) 
      & .478($\uparrow${1.70\%}) 
      & .674($\downarrow${1.03\%})
      \\
      BERT$_{PKGM-R}$ 
      & \textbf{.716}($\uparrow${0.85\%}) 
      & .854($\uparrow${0.59\%}) 
      & .929($\uparrow${0.43\%})
      
      & .615 ($\uparrow${0.82\%})
      & \textbf{.801}($\uparrow${1.39\%})
      &\textbf{.904} ($\uparrow${1.35\%})
      
      & .315($\uparrow${3.96\%}) 
      & \textbf{.507}($\uparrow${7.87\%}) 
      & .691($\uparrow${1.47\%})
      \\
      BERT$_{PKGM-all}$  
      & \textbf{.716}($\uparrow${0.85\%}) 
      & \textbf{.859}($\uparrow${1.18\%}) 
      & \textbf{.932}($\uparrow${0.76\%})
      
      & .615($\uparrow${0.82\%})
      & .799($\uparrow${1.14\%}) 
      & \textbf{.904}($\uparrow${1.35\%})
      
      & \textbf{.321}($\uparrow${5.94\%}) 
      & .502($\uparrow${6.81\%}) 
      & \textbf{.701}($\uparrow${2.94\%})
      \\
    \cline{2-10}
      Average Improvement 
      & \multicolumn{3}{c|}{ $\uparrow${0.754\%}} 
      & \multicolumn{3}{c|}{ $\uparrow${1.003\%}} 
      & \multicolumn{3}{c}{ $\uparrow${3.442\%}} 
      \\
      \bottomrule
    \end{tabular}
    }
    \label{res:classification}
\end{table*}

\subsubsection{Model}
\textbf{Base Model.}
Text classification is an import task in Natural Language Processing(NLP) and text mining, and many methods have been proposed for it. In recent years, deep learning models has been shown outperform traditional classification methods\cite{DBLP:conf/acl/HenaoLCSSWWMZ18, BERT}. Given the input text, the mapping functions $f$ first learns a dense representation of it via representation learning and then uses this representation to perform final classification. 
Recently, large-scale pre-trained language such as ELMo\cite{ELMo}, GPT\cite{GPT} and BERT\cite{BERT}, has become the \emph{de facto} first representation learning step for many NLP tasks. 
We applied BERT as the base model for item classification. 

BERT\cite{BERT} is a pre-trained language model with bidirectional encoder representations from multi-layers of Transformers\cite{transformer}. It pre-trains deep bidirectional representations from the unlabeled text by jointly conditioning on both left and right context in all layers. It is trained on a huge amount of texts with masked language model objective. Since the usage of BERT has become common step of text encoding, we conduct product classification experiment with released pre-trained BERT by Google. We will omit an exhaustive background description of the model architecture and refer the readers to \cite{BERT} as well as the excellent guides and codes\footnote{https://github.com/google-research/bert} of fine-tuning BERT on downstream tasks.

\begin{figure}[!htbp]
     \centering
     \subfigure[Base model.]{
         \centering
         \includegraphics[width=0.202\textwidth]{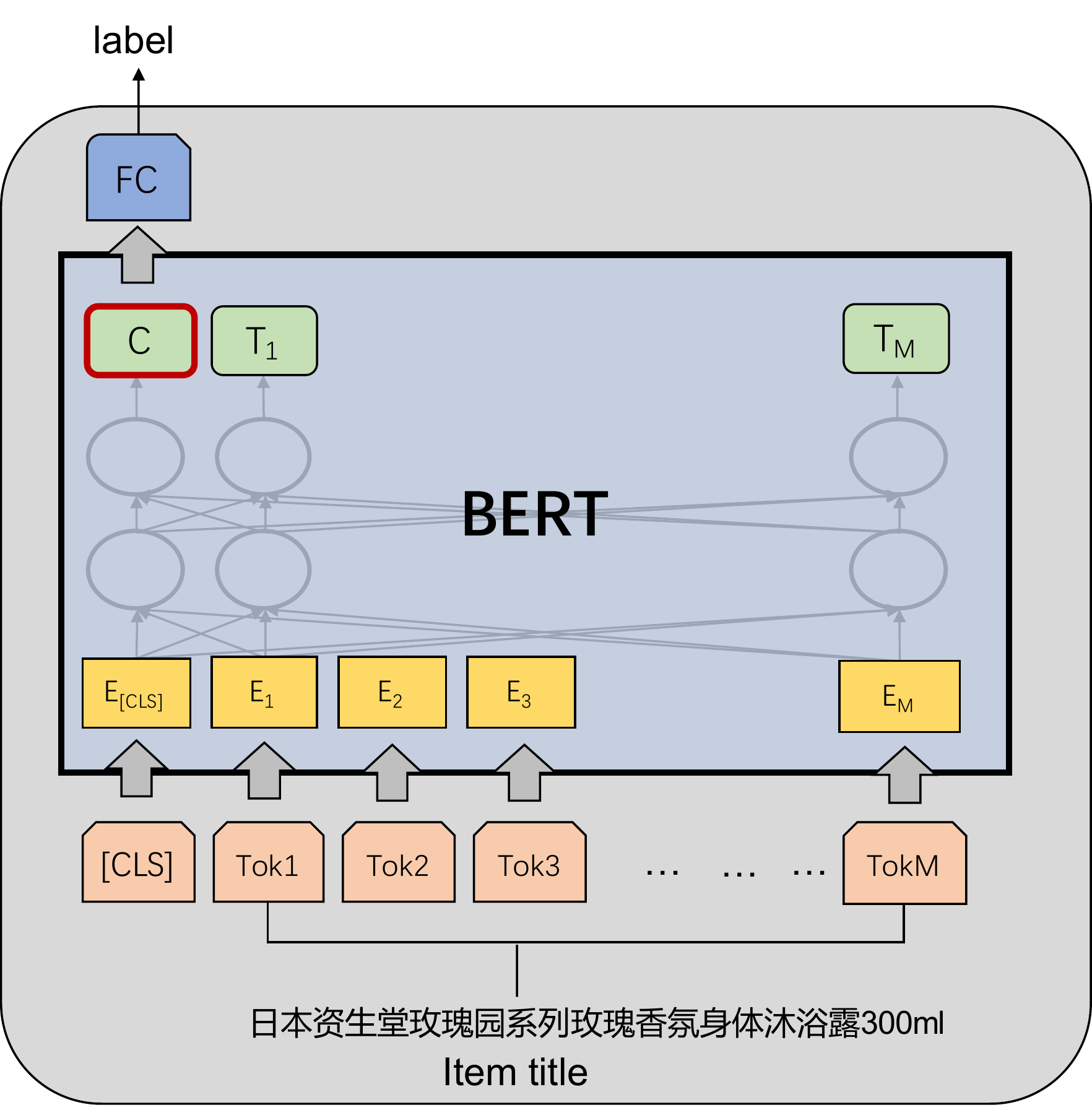}
         \label{fig:cate-base}
     }
     \subfigure[Base model with PKGM services.]{
         \centering
         \includegraphics[width=0.20\textwidth]{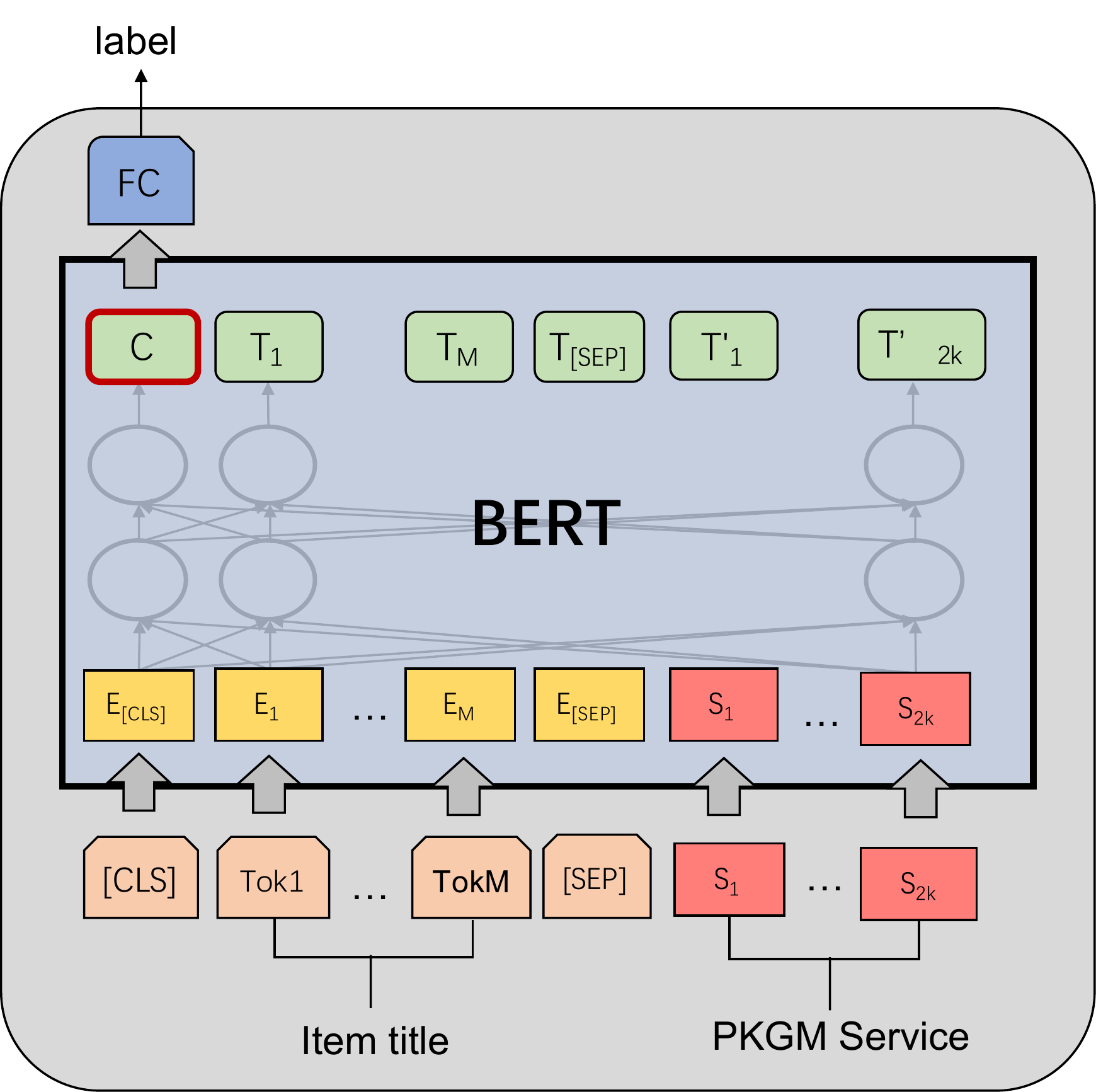}
         \label{fig:cate-base+PKGM}
     }
    \caption{Models for item classification.}
    \label{fig:model-classification}
\end{figure}

We show the details of applying BERT on the item classification task in Figure~\ref{fig:model-classification}. We input  title into BERT, and take out the representation corresponds to [CLS] symbol $C$ for classification with a fully-connected layer as follows
\begin{equation}
    y = \sigma(\mathbf{W}C + \mathbf{b})
\end{equation}
where $\mathbf{W} \in \mathbb{R}^{d \times n_c}$ is a weighed matrix. $d$ is the word embedding dimension, also called hidden size, of BERT and $n_c$ is the number of classes in current task. $\mathbf{b}\in \mathbb{R}^{d}$ is a bias vector and $\sigma(x)$ is the activate function. 

We assign a unique identity number to each class beginning with $0$. And for a product $p \in c_i$ in the dataset, we assign it a label $l_p \in \mathbb{R} ^{n_c}$ which is a one hot vector with $l[i] =1$ and others $0$. 
With predicted $y$ and label $l$ for input title, we fine-tune the model with a cross-entropy loss.

\noindent \textbf{Base+PKGM Model.}
To enhance item classification with our 
PKGM, for each item, 
we provide service vectors or item embedding following Section 4.2. Suppose there are $m$ vectors from PKGM, we first add a [SEP] symbol embedding ahead of these vectors, and then  replace the last $m+1$ tokens of fixed length title sequence with these vectors, as shown in Figure~\ref{fig:model-classification}.
More specifically, for BERT$_{PKGM-all}$, we replace the last $2\times k$ embeddings with full service vectors from PKGM, the  last $k$ embeddings with service vectors from triple query module or relation query module, corresponding to BERT$_{PKGM-T}$ and BERT$_{PKGM-R}$ respectively, and the last one embedding with item embedding for BERT$_{PKGM-item}$.

\subsubsection{Experiment details}
To show the power of pre-trained models, both language model and knowledge graph model, with which we could get good performance on downstream tasks with a few training samples, we experiment on three datasets with number of instance in each class less than 100, 50 and 20, marked as ItemCla-large, ItemCla-medium and ItemCla-small. The ratio of positive and negative samples is $1:1$. Details are shown in Table~\ref{data:classification}.

\begin{table}[!hbpt]
    \centering
    \caption{Data statistic for  item classification task}
    \begin{tabular}{l |c | c | c |c}
    \toprule
         \textbf{Dataset} & \textbf{\# Category} & \textbf{\# Train} & \textbf{\# Test} & \textbf{\# Dev} \\
         \midrule
        ItemCla-large  &1293  & 169039 & 36225 & 36223  \\
        ItemCla-medium & 602 & 20977 & 4497 & 4496\\
        ItemCla-small & 575 & 8398 & 1802 & 1800 \\
        \bottomrule
    \end{tabular}
    \label{data:classification}
\end{table}

We take the pre-trained BERT$_{BASE}$ model trained on Chinese simplified and traditional corpus\footnote{Released at https://storage.googleapis.com/bert\_models/2018\_11\\
\_03/chinese\_L-12\_H-768\_A-12.zip. In this model, number of layers is $12$, the hidden size is $768$ and the number of attention heads is $12$.}. 
Similar to the input format of BERT model, we add a special classification token [CLS] ahead of each sequence,  
which is used as the aggregated sequence representation in the final hidden state. 
In this experiment, we set the length of the sequence to 128\footnote{For titles shorten than $127$ (with [CLS] symbol excepted), we padding it with zero embeddings 
as did in BERT, while for titles longer than $247$, we reserve the first $127$ words.}. 
Finally, we fine-tune BERT with 3 epochs with batch size as 32 and learning rate as $2e^{-5}$.


To evaluate the performance, 
we report $Hit@k(k=1,3,10)$ to show the validity of our method.
$Hit@k$ is calculated by firstly getting the rank of the correct label as its predicted category rank and $Hit@k$ is the percentage of test samples with prediction rank within $k$. 


\subsubsection{Results}
Results in Table~\ref{res:classification} show that BERT$_{PKGM}$ model outperforms BERT model on all datasets with all metrics.
The best result of each metric on all datasets is achieved by either BERT$_{PKGM-all}$ or BERT$_{PKGM-R}$, 
demonstrating the effectiveness of service vectors from PKGM, among which service vectors from relation query module are more informative than those from triple query module for item classification task, which might because properties for items are more informative than concrete values of properties for item classification task. Among three dataset, PKGM helps more on sparser ones.

\subsection{TASK2:Item Resolution}

\subsubsection{Task definition}
The target of item resolution is to find two different items that are referred to  the same product. 
For example,  
there are a lot of IPhoneXI with Green color and 256 GB capacity sold by different online shops, which are stored as different items on the platform, while from the perspective of the product, they refer to  same one. 
With item title as input, item resolution could be framed as a paraphrase identification task\cite{PI}.

\begin{defn}(Item Resolution)
Given a set of data $\mathcal{D} = \{ \mathcal{T},\mathcal{R}, \mathcal{L} \}$.
$\mathcal{T}$ is a set of titles and label set  $\mathcal{L} = \{ True, False\}$. 
$\mathcal{R} = \{(t_1, t_2, l) | t_1 \in \mathcal{T}, t_2 \in \mathcal{T}, l \in \mathcal{L} \} $ is a set of records. $l$ is the label for $t_1$ and $t_2$ referring whether they refer to the same product. 
Title $t = [ w_1, w_2, w_3, ..., w_n ]$ is an ordered sequence composed by words.  The target is to train a mapping function $f: \mathcal{R} \mapsto \mathcal{L}$. 
\end{defn}
\subsubsection{Model}
\textbf{Base Model.} 
Many methods\cite{DBLP:conf/naacl/HeL16, DBLP:conf/naacl/LanX18, BERT}  have been proposed for paraphrase identification in recent years,  
among which the pre-trained language model is state-of-the-art. Thus similar to item classification task, BERT is applied as base model in this task. 



\begin{figure}[!htbp]
     \centering
         \includegraphics[width=0.45\textwidth]{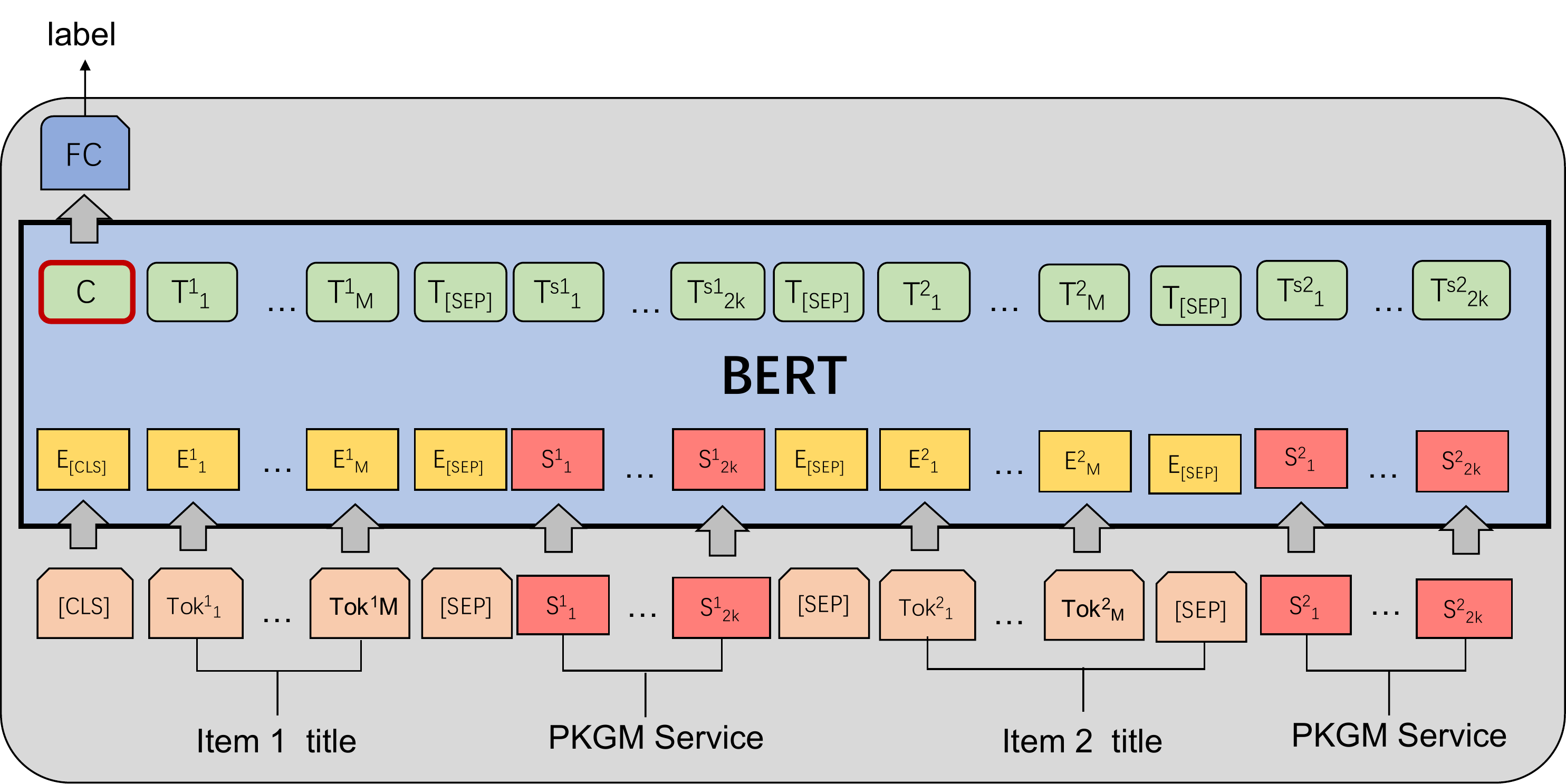}
    \caption{Models for item resolution.}
    \label{fig:model-same}
\end{figure}

\noindent \textbf{Base+PKGM Model.}
We show details of applying BERT supported by PKGM for item resolution in Figure~\ref{fig:model-same}. 
Titles of two items are input and [CLS] symbol $C$ will be used for binary classification as the input for a fully-connected layer(FC)\footnote{
$y = \sigma(\mathbf{W}C + \mathbf{b})$ where $\mathbf{W} \in \mathbb{R} ^{ 2 \times d}$ is a weighted matrix, $d$ is the word embedding dimension, also called hidden size, of BERT,  $\mathbf{b} \in \mathbb{R}^2$ is a bias vector and $\sigma (x)$ is a activation function.}.
We add a [SEP] symbol at the begining of $m$ service vectors from PKGM and then replace the last $m+1$ embeddings of the title with it. 
The model is fine-tuned with cross-entropy loss.


\subsubsection{Experiment details}
We experiment on $3$ datasets created from different categories with different scales. 
Statistic details are shown in Table~\ref{tab:data-same-item}. 

\begin{table}[!hbpt]
    \centering
    \caption{Statistics for item resolution task.}
    \begin{tabular}{l |c | c | c }
    \toprule
        \textbf{Dataset}  & \textbf{\# Train} & \textbf{\# Test} & \textbf{\# Dev} \\
        \midrule
        Girls' Skirts (GS) 
        & 4731 & 1014 & 1013 \\
        Children's Socks (CS)
        & 3968 & 852 & 850 \\
        Hair Decorations (HD)
        & 2424 & 520 & 519 \\
        \bottomrule
    \end{tabular}
    \label{tab:data-same-item}
\end{table}


In the experiment, pre-trained BERT$_{BASE}$\footnote{Released at https://storage.googleapis.com/bert\_models/2018\_11 \\
\_03/chinese\_L-12\_H-768\_A-12.zip} is similar to item classification task with sequence length as 128. 
Each item tile sequence length are adapted to 63  following the same strategies in item classification task. 
We fine-tune all methods for 3 epochs with batch size as 32 and learning rate as $2e^{-5}$.

\subsubsection{Results}

Table~\ref{tab:res-same-item-C} shows the accuracy results for the item resolution. 
Obviously, BERT$_{PKGM-all}$ performs the best on all the datasets. 
BERT$_{PKGM-T}$ and  BERT$_{PKGM-R}$ comparably perform better than baseline BERT$_{PKGM-item}$ and BERT. 
This convincingly demonstrates service vectors from PKGM  promote results of item resolution. 
Improvements on dataset CS are the most, which is significantly larger than dataset GS, showing PKGM helps more for task with sparse datasets. Even dataset HD is sparser than CS, while it is not a harder task then HD and BERT could performs good on it thus PKGM doesn't help much. Thus we could conclude that to a certain extent, the title text with enough training examples could enable BERT good at capture item information, and PKGM helps more with sparse and hard task.

\begin{table}[!hbpt]
    \centering
    \caption{Accuracy results for item resolution.}
    \begin{tabular}{l|c c c }
        \toprule
          & GS & CS  & HD \\
        \midrule
         BERT& .889(0.00\%)  & .869(0.00\%) & .893(0.00\%) \\
         BERT$_{PKGM-item}$ 
         & .889(0.00\%) 
         & .871 ($\uparrow${0.23\%})
         & .894 ($\uparrow${0.11\%})
         \\
         \midrule
         BERT$_{PKGM-T}$ 
         & .887($\downarrow${0.22\%}) 
         & .879($\uparrow${1.15\%}) 
         & .899($\uparrow${0.67\%})
         \\
         BERT$_{PKGM-R}$ 
         & \textbf{.891}($\uparrow${0.22\%})
         & .879($\uparrow${1.15\%}) 
         & .896($\uparrow${0.34\%})
         \\
         BERT$_{PKGM-all}$
         & \textbf{.891}($\uparrow${0.22\%})
         & \textbf{.881}($\uparrow${1.38\%}) 
         & \textbf{.901}($\uparrow${0.90\%})
         \\
        \cline{2-4}
         \textit{Average Improvement} 
         & $\uparrow${0.073\%}
         & $\uparrow${1.227\%} 
         & $\uparrow${0.637\%}
         \\
        \bottomrule
    \end{tabular}
    \label{tab:res-same-item-C}
\end{table}

\subsection{Item Recommendation}
\begin{table*}[!hbpt]
    \centering
    \caption{Results for item recommendation task.}
    \resizebox{\textwidth}{!}{
    \begin{tabular}{l|c|c|c|c|c|c|c|c|c}
    \toprule
      \multirow{2}{*}{\textbf{Method}} & \multicolumn{3}{c|}{\textbf{TAOBAO-Recom-large}}  & \multicolumn{3}{c|}{\textbf{TAOBAO-Recom-medium}}  & \multicolumn{3}{c}{\textbf{TAOBAO-Recom-small}}  \\
      \cline{2-10}
         & NDCG@5 &NDCG@10 & NDCG@30 & NDCG@5 &NDCG@10 & NDCG@30 & NDCG@5 &NDCG@10 & NDCG@30\\
         \midrule
        NCF  
        & .407(0.00\%) & .442(0.00\%) & .485(0.00\%) 
        & .360(0.00\%) & .392(0.00\%) & .435(0.00\%) 
        & .350(0.00\%) & .382(0.00\%) & .425(0.00\%) \\
        NCF$_{PKGM-item}$ 
        & .407 (\textcolor{black}{$0.00\%$}) 
        & .444($\uparrow${$0.45\%$}) 
        & .487($\uparrow${$0.41\%$})
        & .356($\downarrow${$1.11\%$}) 
        & .387($\downarrow${$1.28\%$}) 
        & .431($\downarrow${$0.92\%$}) 
        & .344($\downarrow${$1.71\%$}) 
        & .378($\downarrow${$1.05\%$}) 
        & .422($\downarrow${$0.71\%$})  \\
        \midrule
        NCF$_{PKGM-T}$ 
        & .409($\uparrow${$0.49\%$}) & .445($\uparrow${$0.68\%$}) & .488($\uparrow${$0.62\%$})
        & .368($\uparrow${$2.22\%$}) & .400($\uparrow${$2.04\%$}) & .444($\uparrow${$2.07\%$}) 
        & .358($\uparrow${$2.29\%$}) &.389($\uparrow${$2.62\%$}) 
        & .431 ($\uparrow${$1.41\%$})
        \\
        NCF$_{PKGM-R}$ 
        & \textbf{.442}($\uparrow${$8.60\%$}) & \textbf{.478}($\uparrow${$8.14\%$}) & \textbf{.520}($\uparrow${$7.22\%$})
        & \textbf{.397}($\uparrow${$10.28\%$}) & \textbf{.430}($\uparrow${$9.69\%$}) &\textbf{.475}($\uparrow${$9.20\%$}) 
        & \textbf{.392}($\uparrow${$12.00\%$}) & \textbf{.424}($\uparrow${$10.99\%$}) & \textbf{.466}($\uparrow${$9.65\%$}) \\
        NCF$_{PKGM-all}$ 
        & .440($\uparrow${$8.11\%$}) & .476($\uparrow${$7.69\%$}) & .519($\uparrow${$7.01\%$}) 
        & .390($\uparrow${$8.33\%$}) & .424($\uparrow${$8.16\%$}) & .469($\uparrow${$7.82\%$}) 
        & .380($\uparrow${$8.57\%$}) & .414($\uparrow${$8.38\%$}) & .458($\uparrow${$7.76\%$})\\
        \cline{2-10}
        Average Improvement 
        & \multicolumn{3}{c|}{$\uparrow${5.396\%}} 
        & \multicolumn{3}{c|}{$\uparrow${6.646\%}}
        & \multicolumn{3}{c}{$\uparrow${6.987\%}}\\
        \bottomrule
        
    \end{tabular}
    }
    \label{tab:res-recommendation}
\end{table*}

\subsubsection{Task definition}
Item recommendation aims to properly recommend items to users with a high probability to have interactions. 
To learn user preference and make proper recommendations from observed implicit (click, buy, etc) user feedback, it is formed as a ranking problem \cite{DBLP:conf/www/HeLZNHC17}.

\begin{defn}(Item Recommendation)
Given a set of data $\mathcal{D} = \{\mathcal{U}, \mathcal{I}, \mathcal{R}, \mathcal{S}\}$, where $\mathcal{U}$ and $\mathcal{I}$ are a set of users and items respectively, $\mathcal{R}$ and $\mathcal{S}$ is a set of interactions between users and items and their scores\footnote{Score of each interaction is $1$ by default.}. The target is to train a mapping function $f : \mathcal{R} \mapsto \mathcal{S}$. 
\end{defn}


\subsubsection{Model}

\textbf{Base Model.}
We adopt Neural Collaborative Filtering (NCF)\cite{DBLP:conf/www/HeLZNHC17} as a general framework of our base model. In NCF,  Generalized Matrix Factorization (GMF) and Multi-Layer Perceptron (MLP) are used for modeling user-item interaction, in which GMF uses a linear kernel to model the latent feature interactions, and MLP uses a non-linear kernel to learn the interaction function from data.
In NCF, there are 4 layers including input layer, embedding layer, neural collaborative filtering  layer, and output layer. 
Input layer contains user and item sparse feature vectors $\mathbf{v}_u^U$ and $\mathbf{v}_i^I$ that describe user $u$ and item $i$ respectively, which are set to one-hot sparse vectors in our experiments following  pure collaborative setting.
Embedding layer is a fully-connected network to project one-hot representations to dense embeddings. 
In neural collaborative filtering layer, there are GMF and MLP to model item-use interactions, which will output hidden-state representations. With hidden-state representations, output layer will first concatenate them and then output a score for the input user-item pair to indicating the possibility of interactions between them. Too detailed description of the architecture of NCF is omit and we refer readers to \cite{DBLP:conf/www/HeLZNHC17}.

\begin{figure}[!htbp]
         \centering
         \includegraphics[width=0.45\textwidth]{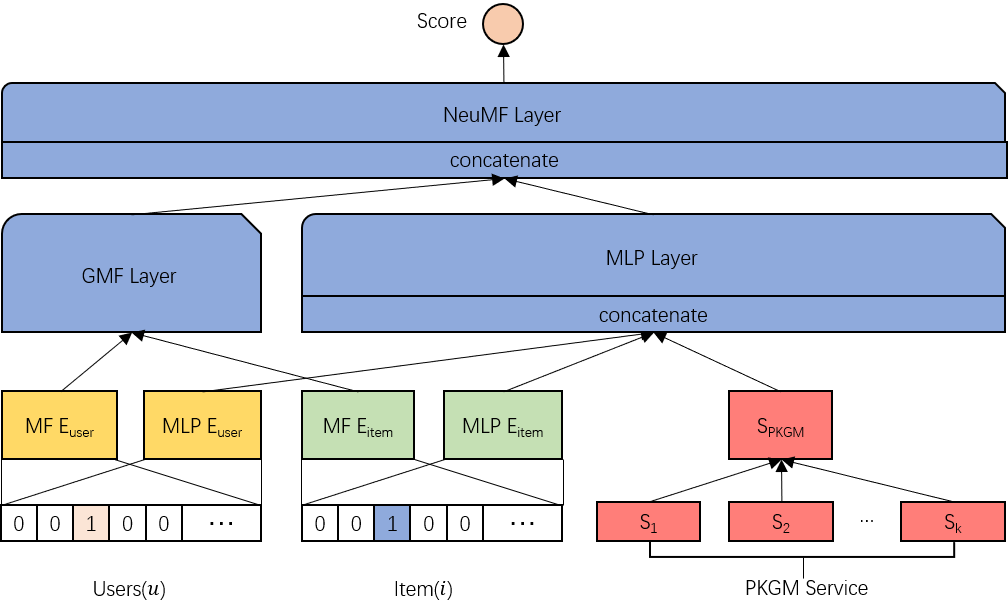}
    \caption{Architecture of NCF$_{PKGM}$ for recommendation.}
    \label{fig:model-recommendation}
\end{figure}

\noindent \textbf{Base+PKGM Model.}
Since NCF is built based on item embeddings, we integrate  services vectors from PKGM with item embeddings in the second way as introduced before. Specifically, for item $e$ in each user-item pair, similar to previous tasks, $2k$ service vectors for each item will be provided by PKGM,  
denoted as $[S_1, S_2, ..., S_{2k}]$. 
We first make them into one:
\begin{equation}
    S^{e} = \frac{1}{k} \sum_{i \in [1, 2, ..., k]} [S_i;S_{i + k}]
\end{equation}
where $[A; B]$ means concatenation of vector $A$ and $B$. Then we integrate $S^e$ into MLP layer by making the vertical concatenation of three vectors as input
\begin{equation}
    \phi ^{MLP} (\mathbf{p}_u, \mathbf{q}_i, S^e) = {
\left[ \begin{array}{ccc}
\mathbf{p}_u \\
\mathbf{q}_i \\
S_{PKGM} 
\end{array} 
\right ]}
\end{equation}
where $\mathbf{p}_u$ and $\mathbf{q}_i$ is the user embedding and item embedding for MLP layer. 
Other parts of NCF stays the same. Fig.~\ref{fig:model-recommendation} shows the general framework of NCF$_{PKGM}$.

\subsubsection{Experiment details}
We conduct experiments on three datasets sampled from records on Taobao platform. Their statistics are shown in TABLE \ref{tab:taobao-recommendation-statistics}.
In each dataset, the number of interactions for each user is ensured to more than $10$. 

\begin{table}[!hbpt]
    \centering
    \caption{Data statistics for item recommendation task.}
    \begin{tabular}{l |c|c | c }
    \toprule
        \textbf{Dataset} & 
        \textbf{\# Items} & \textbf{\# Users} & \textbf{\# Interactions} \\
         \midrule
        TAOBAO-Recom-large  & 37847 & 29015 & 443425 \\
        TAOBAO-Recom-medium & 34135 & 14188 & 217479\\
        TAOBAO-Recom-small & 32449 & 11393 & 174869\\
    \bottomrule
    \end{tabular}
    \label{tab:taobao-recommendation-statistics}
\end{table}

The dimension of user and item embedding in GMF layer are set to $8$ and $32$ for MLP layer. 
Three hidden layers of size $[32, 16, 8]$ after embedding layer are used.
The model is learned with same loss 
proposed in \cite{DBLP:conf/www/HeLZNHC17} 
with an external $L2$ regularization with regularization factor as $ 0.001$ on user and item embedding layer in MLP and GMF. 
Models are optimized by Adam\cite{Adam} with learning rate $0.0001$, and trained for $100$ epochs with bach size of $256$. 
During training, we use a negative sampling ratio of $4$. 

Performance of item recommendation is evaluated by the \textit{leave-one-out} strategy which is used in \cite{DBLP:conf/www/HeLZNHC17}. For each user, we held-out the latest interaction as the test
set and others are used in the train set. As for the testing procedure, we uniformly sampled 100 unobserved negative items, ranking the positive test item with negative ones. 
For evaluation metrics, Normalized Discounted Cumulative Gain (NDCG)@k are used where k =
$[5,10,30]$.


\begin{table*}[!hbpt]
    \centering
    \caption{Accuracy results for scene detection.}
    \begin{tabular}{l|c c|  c  c | c c}
    \toprule
    \textbf{Dataset} & \multicolumn{2}{c|}{\textbf{SceneDetect-large}} &  \multicolumn{2}{c|}{\textbf{SceneDetect-medium}} &  \multicolumn{2}{c}{\textbf{SceneDetect-small}} \\
     \midrule
     \textbf{Method}    & Base=MobileNetV2 & Base=ResNet & Base=MobileNetV2 & Base=ResNet & Base=MobileNetV2 & Base=ResNet\\
        \midrule
        Base 
        & .502(0.00\%) 
        & .499(0.00\%)   
        & .466(0.00\%)
        & .476(0.00\%)
        & .429(0.00\%)
        & .419(0.00\%)
        \\
        
        Base$_{PKGM-item}$ 
        & .501($\downarrow${0.20\%})
        & .502($\uparrow${0.60\%})
        & .463($\downarrow${0.64\%})
        & .473($\downarrow${0.63\%})
        & .433($\uparrow${0.93\%})
        & .414($\downarrow${1.19\%})
        \\
        \midrule
        
        Base$_{PKGM-T}$ 
        & .505($\uparrow${0.60\%})
        & .507($\uparrow${1.60\%}) 
        & .467($\uparrow${0.21\%})
        & .476(\textcolor{black}{0.00\%})  
        & .433($\uparrow${0.93\%})
        & .419(\textcolor{black}{0.00\%})
        \\
        
        Base$_{PKGM-R}$ 
        & .505($\uparrow${0.60\%})
        & .507($\uparrow${1.60\%})
        & .466(\textcolor{black}{0.00\%})
        & .475($\downarrow${0.21\%})
        & .431($\uparrow${0.47\%})
        & \textbf{.423} ($\uparrow${0.95\%})
        \\
        
        Base$_{PKGM-all}$ 
        & \textbf{.508}($\uparrow${1.20\%}) 
        & \textbf{.511}($\uparrow${2.40\%})  
        & \textbf{.472}($\uparrow${1.29\%}) 
        & \textbf{.478}($\uparrow${0.42\%}) 
        & \textbf{.442}($\uparrow${3.03\%})
        & .417($\downarrow${0.48\%})
        \\
        \cline{2-7}
        Average Improvement 
        & \multicolumn{2}{c|}{ $\uparrow${1.333 \%}}
        & \multicolumn{2}{c|}{ $\uparrow${0.285 \%}}
        & \multicolumn{2}{c|}{ $\uparrow${0.817 \%}}\\
        \bottomrule
    \end{tabular}
    \label{tab:res-scene-detection}
\end{table*}

\subsubsection{Results}

Experiment result are shown in TABLE \ref{tab:res-recommendation}. 
We analyse the results as follows. 

Firstly, all of the PKGM enhanced NCF models outperform the base NCF in all metrics. 
 NCF$_{PKGM-T}$,  NCF$_{PKGM-R}$ and  NCF$_{PKGM-all}$
 outperforms the NCF baseline with an average promotion of 
$0.23\%$, $3.43\%$ and $3.43\%$ on NDCG metrics respectively. 
Such promotion proves that service vectors provided  our PKGM successfully provide external information that are not included in original user-item interaction data and they are helpful for recommendation.
Secondly, 
performances of NCF$_{PKGM-R}$ are better than those on NCF$_{PKGM-T}$ and NCF$_{PKGM-all}$. Thus service vectors from relation query module are more useful than the those from triple query module for item recommendation, which is largely due to the fact that properties of items are more effective than detailed values for capturing user preferences.
Thirdly, PKGM helps more for sparse datasets. 
\subsection{Scene Detection}
\subsubsection{Task definition}
The target of scene detection is to detect the usage scene of items, such as hiking, gifts for mothers and growing flowers. Such detection is useful for data management and also recommendation.
Item pictures are usually used for this task, and it could be formulated as picture classification problem. 
\begin{defn}(Scene Detection)
Given a set of data $\mathcal{D} = \{\mathcal{I}, \mathcal{P}, \mathcal{S}, \mathcal{R}\}$, where $\mathcal{I}, \mathcal{P},  \mathcal{S}$ and  $\mathcal{R}$ are a set of items, pictures, scenes and records respectively, and $\mathcal{R} = \{(item, \{p_1, p_2, ... \}, s)| item \in \mathcal{I}, p_i \in \mathcal{P}, s\in \mathcal{S} \}$ and each records includes multiple pictures of one item with its scene label, the task is to learn a mapping function $f : \mathcal{I} \mapsto \mathcal{S}$
\end{defn}

\subsubsection{Model}
\noindent \textbf{Base Model.}
Many methods\cite{ResNet, MobileNetV2, DBLP:conf/cvpr/WangJQYLZWT17} have been proposed for image classification in recent years. For image processing, image encoder is a key part which will effect the final results significantly. Many pre-trained image encoder have been proposed and widely applied in image related tasks, among which we applied two of the most popular models, ResNet\cite{ResNet} and MobileNetV2\cite{MobileNetV2}. 
We pre-train them from scrach on datasets from e-commerce domain to make them better adapt to e-commerce application. The dataset for image pre-training includes more than 90 million images and 6 thousands classes. ResNet is set to $50$ layers and MobileNetV2 to $101$ layers and hidden size is set to $1024$ for both of them.  For one item, there are multiple images, thus we firstly add all image embeddings into one and regard it as item embedding from images, and then input it into a fully-connected layer for classification. 

\noindent \textbf{Base+PKGM Model.}
Both ResNet and MobileNetV2 are based on item embeddings via image encoder, we integrate service vectors from PKGM in the second way as introduced before. Specifically, for each item, $2k$ service vectors $[S_1, S_2, ..., S_{2k}]$ are  provided, and we first make them into one
\begin{equation}
    S_{PKGM} = [S_1 ; S_2 ; ... ;S_{2k}]
\end{equation}
where $[A;B]$ indicates concatenation of A and B. Then we integrate $S_{PKGM}$ with item embedding via concatenation $[E; S_{PKGM}]$ which will replace $E$ as the input for final fully-connected classification layer. Both base model and PKGM enhanced model are trained with cross entropy loss.  

\begin{figure}[!htbp]
         \centering
         \includegraphics[width=0.45\textwidth]{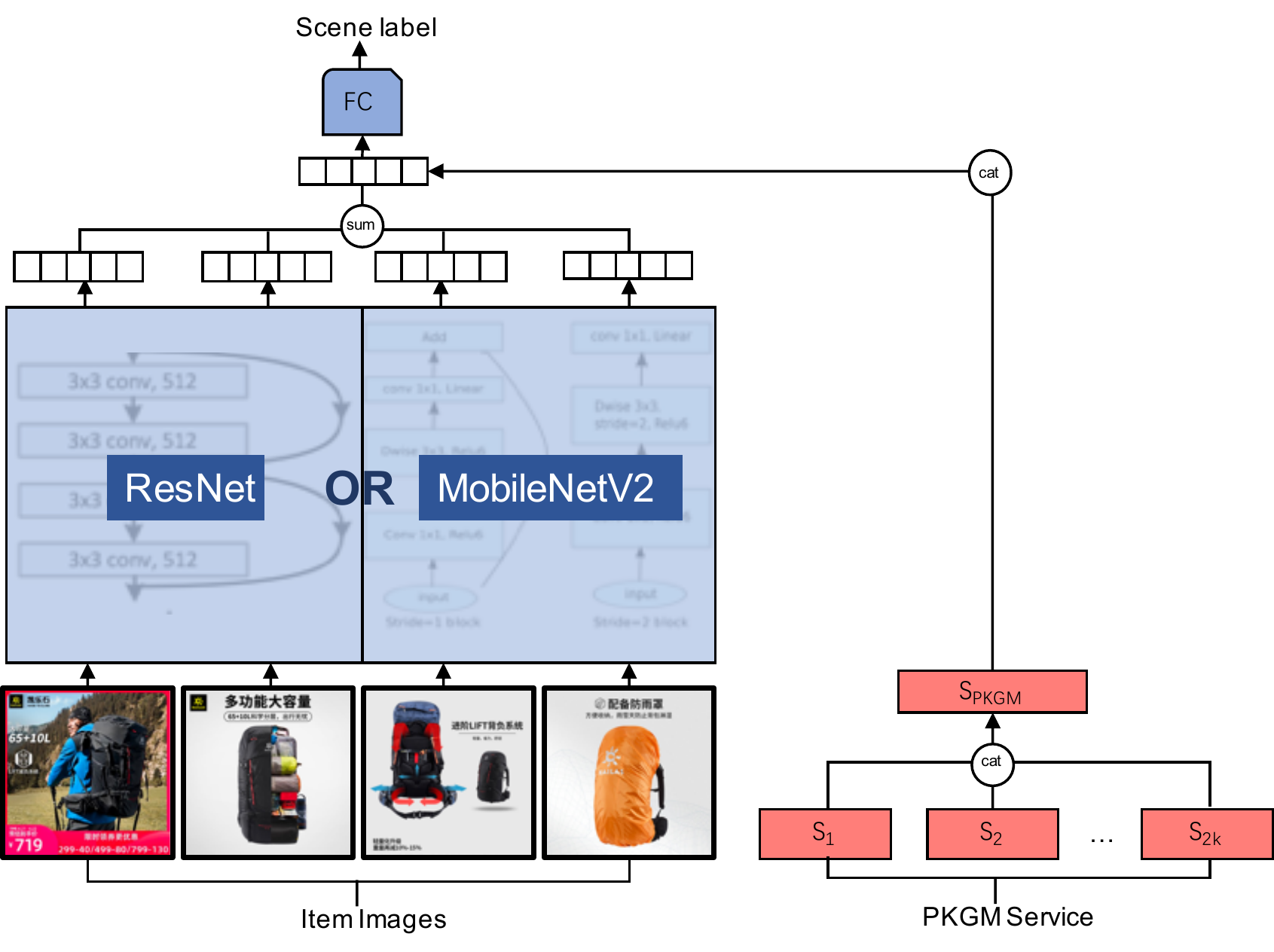}
    \caption{Architecture of ResNet/MobileNetV2$_{PKGM}$.}
    \label{fig:model-scene}
\end{figure}


\begin{table}[!hbpt]
    \centering
    \caption{Statistic of datasets for scene detection task.}
    \begin{tabular}{l|c| c |c }
        \toprule
         \multirow{2}{*}{\textbf{Dataset}} & \multirow{2}{*}{\textbf{\# Classes}}  & \multicolumn{2}{c}{\textbf{\# Items} / \textbf{\# Images}} \\
          \cline{3-4}
          & & Train &  Test \\
        \midrule
         SceneDetect-large &132 &22524/90096 &5632/22520 \\
         SceneDetect-medium &134 &2125/8500 & 532/2128 \\
         SceneDetect-small &41 &205/820 & 52/208 \\
        \bottomrule
    \end{tabular}
    \label{tab:data-scene-detection}
\end{table}

\begin{table*}[!hbpt]
    \centering
    \caption{Results for sequential recommendation task.}
    \resizebox{\textwidth}{!}{
    \begin{tabular}{l|c|c|c |c|c|c |c|c|c}
    \toprule
        \multirow{2}{*}{\textbf{Method}} & \multicolumn{3}{c|}{\textbf{SeqRec-large}} & \multicolumn{3}{c|}{\textbf{SeqRec-medium}} & \multicolumn{3}{c}{\textbf{SeqRec-small}} \\  \cline{2-10}
        & Hit@10 & NDCG@10 & MRR  & Hit@10 & NDCG@10 & MRR & Hit@10 & NDCG@10 & MRR\\
        \midrule
        Fissa 
        & .792(0.00\%) &.513(0.00\%) &.425(0.00\%)
        & .790(0.00\%) & .509(0.00\%) &.420(0.00\%) 
        & .778(0.00\%) & .501(0.00\%) & .414(0.00\%) \\
        
        Fissa$_{PKGM-item}$ 
        & .804($\uparrow${$1.52\%$}) 
        & .534($\uparrow${$4.09\%$}) 
        & .449($\uparrow${$5.65\%$})
        & .799($\uparrow${$1.14\%$}) 
        & .529($\uparrow${$3.93\%$}) 
        & .443($\uparrow${$5.48\%$}) 
        & .794($\uparrow${$2.06\%$}) 
        & .516($\uparrow${$2.99\%$}) 
        & .429($\uparrow${$3.62\%$}) \\
        \midrule
        
        Fissa$_{PKGM-T}$ 
        & \textbf{.819}($\uparrow${$3.41\%$}) 
        & \textbf{.544}($\uparrow${$6.04\%$}) 
        & .457($\uparrow${$7.53\%$}) 
        & .816($\uparrow${$3.29\%$}) 
        & .540($\uparrow${$6.09\%$}) 
        & .453($\uparrow${$7.86\%$}) 
        & .812($\uparrow${$4.37\%$}) 
        & .539($\uparrow${$7.58\%$}) 
        & .453($\uparrow${$9.42\%$}) \\
        
        Fissa$_{PKGM-R}$ 
        & .812($\uparrow${$2.53\%$}) 
        & .542($\uparrow${$5.65\%$}) 
        & \textbf{.457}($\uparrow${$7.53\%$}) 
        & .807($\uparrow${$2.15\%$}) 
        & .537($\uparrow${$5.50\%$}) 
        & .451($\uparrow${$7.38\%$}) 
        & .799($\uparrow${$2.70\%$}) 
        & .533($\uparrow${$6.39\%$}) 
        & .449($\uparrow${$8.45\%$}) \\
        Fissa$_{PKGM-all}$ 
        & .798($\uparrow${$0.76\%$}) 
        & .532($\uparrow${$3.70\%$}) 
        & .448($\uparrow${$5.41\%$}) 
        & .797($\uparrow${$0.89\%$}) 
        & .536($\uparrow${$5.30\%$}) 
        & .453($\uparrow${$7.86\%$}) 
        & .801($\uparrow${$2.96\%$}) 
        & .531($\uparrow${$5.99\%$}) 
        & .446($\uparrow${$7.73\%$}) \\
        \cline{2-10}
        Average Improvement 
        & \multicolumn{3}{c|}{$\uparrow${4.729 \%}} 
        & \multicolumn{3}{c|}{$\uparrow${5.147 \%} }
        & \multicolumn{3}{c|}{$\uparrow${6.177 \%}} \\
        \bottomrule
    \end{tabular}
    }
    \label{tab:res-sequential-recommendation}
\end{table*}

\subsubsection{Experiment details}
Scene detection experiments are conduct on three datasets with different scales shown in  TABLE~\ref{tab:data-scene-detection}. 
In each dataset, one item has averaged four related images.
Both Base model and the Base+PKGM model are trained with the same hyperparameters. The adam optimizer\cite{Adam} with a learning rate of 0.001 is used, and 1000 epochs are trained with a batch size of 1024. 
\subsubsection{Results}
The average accuracy results of 10 runs are reported in TABLE~\ref{tab:res-scene-detection}. Results show that enhanced with service vectors from PKGM, the performance of base models are improved over three datasets, and the improvement ratio significantly outperform baseline Base$_{PKGM-item}$, showing the effectiveness of service vectors. The largest and smallest improvements show in the large and medium dataset, illustrating that the how helpful PKGM is will also be influenced by the base model's performance and sparsity of datasets. 

\subsection{Sequential Recommendation}
\subsubsection{Task definition}

\begin{defn}(Sequential Recommendation)
Given user set $\mathcal{U}$ and item set $\mathcal{I}$, for each of the user $u \in \mathcal{U}$, its interaction history is defined as $S^u = \{s_{1}^u, s_{2}^u, ..., s_{|S_u|}^u\}$ where $s_{i}^u \in \mathcal{I}$. The target of the sequential recommendation is to predict an  item list in which real next interaction $s^u_{|S_u|+1} \in \mathcal{I} \backslash S^u$ ranked highly.
\end{defn}
\subsubsection{Model}

\noindent \textbf{Base Model.}
We apply a recently proposed global-local representation mixing attention-aware model FISSA as base model, which consists of three module: 1) local representation learning, 2) global representation learning and 3) gating. FISSA treat the state-of-the art self-attentive sequential recommendation (SASRec) model as the local representation learning module to capture the dynamic preference beneath user's behavior sequences. The global representation learnign module is devised to improve the modeling of user's global preference. The gating module is used to balances the local and global representations by taking the information of the candidate items into account. The final representation $z_l$ is a mixture of local and global representation. And the prediction of probability for item $i$ to be the $l$th item is 
\begin{equation}
    r_{\ell+1, i}=z_{\ell}\left(m_{i}\right)^{T}
\end{equation}
where $m_{i}$ is the candidate item embedding.

\begin{figure}[!htbp]
         \centering
         \includegraphics[width=0.49\textwidth]{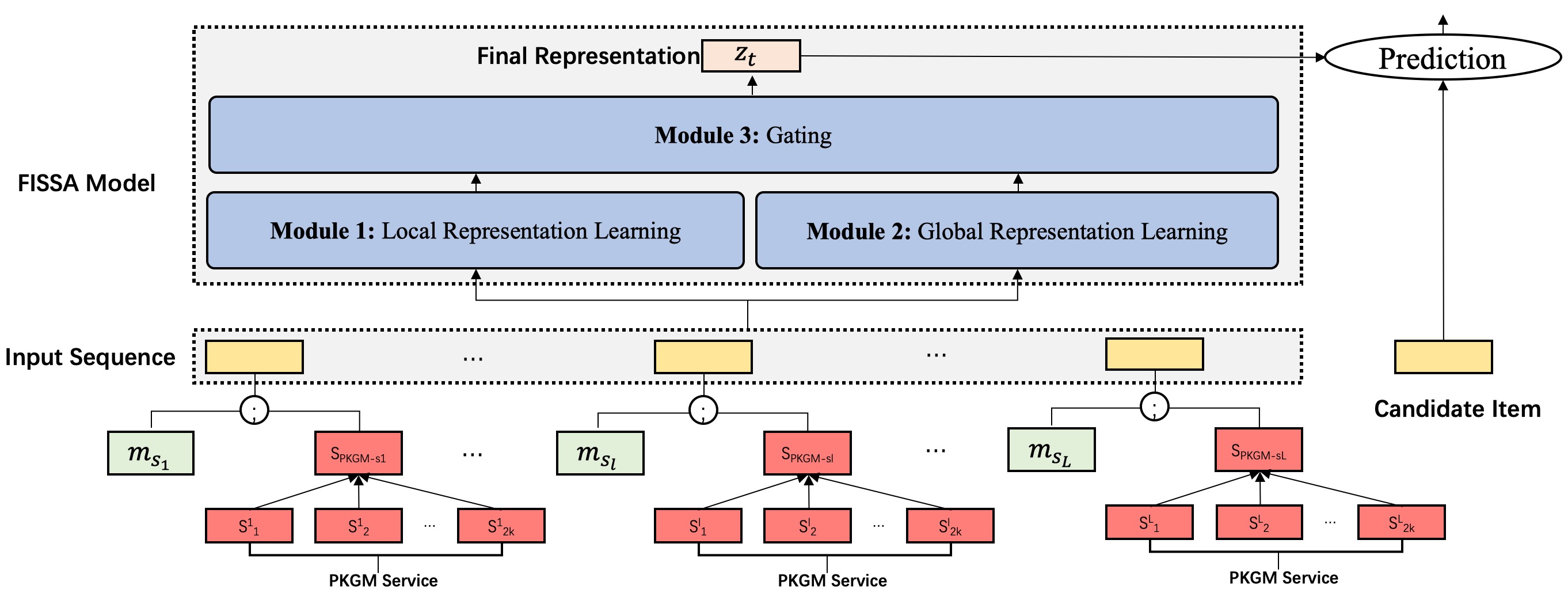}
    \caption{Architecture of FISSA$_{PKGM}$.}
    \label{fig:model-sequential}
\end{figure}

\noindent \textbf{Base+PKGM Model.}
In base model, item embedding matrix $E$ is used in local representation learning procedure as well as global representation learning procedure. We integrate service vectors from PKGM with the learnable embedding matrix $E$ by concatenation
\begin{equation}
    E_{new} = [E;S_{PKGM}]
\end{equation}
where [;] represents concatenating operation. 
And the structure of Base+PKGM model is illustrated in the figure\ref{fig:model-sequential}.
\subsubsection{Experiment details}
Sequential recommendation experiments are conduct on three datasets sampled from Taobao platform in different scale. In each dataset, users and items all have more than 5 interactions. TABLE~\ref{tab:data-sequential-recommendation} shows their details.

For both base model and base+PKGM model, we use the same hyperparameter settings. Models are optimized by Adam\cite{Adam} with learning rating 0.001, and trained for 1000 epochs with a batch size of 128. The maximum length per user sequence is 50 following FISSA setting.
\subsubsection{Results}
The results of sequential recommendation is illustrated in TABLE~\ref{tab:res-sequential-recommendation}. 

Fitstly,  FISSA$_{PKGM}$ significantly improve FISSA over three datasets and FISSA$_{PKGM-T}$ achieves the best performance, confirming the usability of PKGM model. 
Secondly, we observe that with decrease of dataset size, the average percentage of improvement among all of three evaluation metrics increases, which probably due to the decrease of average session length of three different datasets. Thus PKGM model can promote fitness of original model for more sparse data.
Thirdly, PKGM helps more for tasks with sparser dataset in sequential recommendation task. 

\begin{table}[]
    \centering
    \caption{Statistic of dataset for sequential recommendation task.}
    \begin{tabular}{l|c c c }
        \toprule
         \textbf{Dataset} & \textbf{\# Avg session length} & \textbf{\# User} & \textbf{\# Item} \\
        \midrule
         SeqRec-large & 6.36 & 8288 & 8068\\
         SeqRec-medium & 6.36 & 5552 & 7738\\
         SeqRec-small & 6.33 & 4144 & 7265 \\
        \bottomrule
    \end{tabular}
    \label{tab:data-sequential-recommendation}
\end{table}

\section{Related work}

\textbf{Knowledge Graph Embedding (KGE).}
KGE methods learn representations for entities and relations that preserve the information contained in the graph. KGEs have been applied to various tasks including link prediction, entity classification, and entity alignment. 
TransE\cite{TransE-Bordes-2013} is the first KGE model that propose to embed entities and relation in a triple $(h,r,t)$ into continuous vector space, assuming that $\mathbf{h}+\mathbf{r}= \mathbf{t}$. Following TransE, many methods~\cite{TransH-Wang-2014,TransR-Lin-2015,TransD-Ji-2015,TranSparse-Ji-2016} 
are proposed to improve TransE of encoding N-to-N relations, multiple semantics of relations and so on. 
Apart from translation-based models following TransE in real-value space,  semantic matching vector space assumption~\cite{RESCAL-Nickel-2011,DistMult-Yang-2015}, complex-value space~\cite{ComplEx-Trouillon-2016}, hyperbolic space~\cite{DBLP:conf/nips/BalazevicAH19, DBLP:conf/acl/ChamiWJSRR20} and convolutional neural networks~\cite{ConvE-Dettmers-2018} are also introduced. 



\textbf{Pre-trained Language Models with Knowledge.}
The pre-trained language models (PLMs)~\cite{BERT,XLNet-Yang-2019,gpt3_brown2020language} have achieved excellent performance in many NLP tasks and brought convenience in completing various specific downstream tasks,
which greatly inspire us to pre-train a knowledge graph.
Based on PLMs, there is a line of work to inject knowledge from KG, to enable PLMs to gain better performance with the awareness of knowledge.  
There are mainly  XX kinds of ways to inject triples. The first one is reordering input sentence and generate knowledge mask used to attention maps\cite{K_BERT-Liu-2019}. 
The second one is integrating informative entity representation into PLMs via static or dynamic aligning and aggregating method and devise entity specific pretrain tasks~\cite{ERNIE-Zhang-2019, KnowBert-Peters-2019,su2020DKPLM}. 
The third one is to align entities with and inject triples into language model via encoding descriptions of entities with the same encoder with input text~\cite{KEPLER-Wang-2019,yu2020JAKET}.  
The fourth one is to conduct the second-round pretraining based on PLMs with specifically designed knowledge-enriched tasks, where factual knowledge is first reformed into sentence~\cite{wang2020K_Adapter}. 
Also, injecting knowledge into PLM for different application tasks are also widely researched\cite{rosset2020KALM,xiong2020progressively,lauscher2019informing}. 

Different from injecting knowledge into PLMs, we explore how to pretrain knowledge graph and provide knowledge service for various downstream tasks not only including NLP tasks but also other tasks, such as recommendation and image classification.

\section{Conclusion}
In this work, we present our experience of pre-training knowledge graph to provide knowledge services for other tasks with billion-scale product knowledge graph in e-commerce application. We propose Pre-trained Knowledge Graph Mode (PKGM) with two modules to handle triple queries and relation queries in vector space, which could provide  item knowledge services in a uniform way for embedding-based models without accessing triple data. We also propose general ways of applying service vectors in different downstream task models. 

With five types of downstream tasks, we prove that 
(1) PKGM contain rich knowledge about items since with services from PKGM, the performance of item-knowledge-related tasks are successfully improved; 
(2) compared to help with item embeddings, service vectors from triple and relation query module are more helpful for downstream tasks
(3) generally PKGM is more helpful for item-knowledge-related tasks with sparse dataset, but it also depends on the original performance of base models. 

In future work, 
(1) firstly, we would like to explore the potential of PKGM in both e-commerce and open-domains such as applying PKGM to more diverse item-knowledge-related tasks and adopt it to pre-train open-domain knowledge graphs such Wikidata and DBpedia. 
(2) Secondly, we would like to adapt PKGM to more widely used and effective knowledge graph embedding methods such as DistMult and ConvE. 
(3) Thirdly, apart from triple and relation query module, we also would like to explore other potential and important query modules could be applied into PKGM to make the knowledge in service vectors more diverse. 



%



\ifCLASSOPTIONcompsoc
  \section*{Acknowledgments}
\else
  \section*{Acknowledgment}
\fi

This work is funded by NSFCU19B2027/91846204/61473260, national key research program 2018YFB1402800 and supported by Alibaba Group through Alibaba Innovative Research Program.

\ifCLASSOPTIONcaptionsoff
  \newpage
\fi



%



\bibliographystyle{IEEEtran}
\bibliography{pretrain}

\end{document}